\documentclass[pdflatex,sn-mathphys-num]{sn-jnl}

\usepackage{graphicx}%
\usepackage{multirow}%
\usepackage{amsmath,amssymb,amsfonts}%
\usepackage{amsthm}%
\usepackage{mathrsfs}%
\usepackage[title]{appendix}%
\usepackage{xcolor}%
\usepackage[most]{tcolorbox}
\usepackage{textcomp}%
\usepackage{manyfoot}%
\usepackage{siunitx}
\usepackage{booktabs}%
\usepackage{algorithm}%
\usepackage{algorithmicx}%
\usepackage{algpseudocode}%
\usepackage{listings}%
\usepackage{caption}
\usepackage{subcaption}
\usepackage[normalem]{ulem}

\usepackage{arabtex}
\usepackage{utf8}

\theoremstyle{thmstyleone}%
\theoremstyle{thmstyletwo}%

\theoremstyle{thmstylethree}%

\newtcolorbox
  [auto counter, number within=section] 
  {promptbox}[2][]{%
    colback=gray!10,
    colframe=gray!50,
    boxrule=0.5pt,
    arc=2pt,
    left=6pt, right=6pt, top=6pt, bottom=6pt,
    fonttitle=\bfseries,
    title={Prompt~\thetcbcounter: #2},
    #1
}

\begin{document}
\setcode{utf8}

\title[Article Title]{AraTable: Benchmarking LLMs' Reasoning and Understanding of Arabic Tabular Data}


\author[1]{\fnm{Rana} \sur{Alshaikh}}\email{raalshaikh@kau.edu.sa}

\author[2]{\fnm{Israa} \sur{Alghanmi}}\email{iaalghanmi@uj.edu.sa}

\author*[3]{\fnm{Shelan} \sur{Jeawak}}\email{shelan.jeawak@uwe.ac.uk}

\affil[1]{\orgdiv{Department of Information Systems, Faculty of Computing and Information Technology}, \orgname{King Abdulaziz University}, \orgaddress{ \city{Rabigh}, \postcode{21911}, \country{Saudi Arabia}} }

\affil[2]{\orgdiv{Department of Information Systems and Technology, College of Computer Science and Engineering}, \orgname{University of Jeddah}, \orgaddress{ \city{Jeddah}, \country{Saudi Arabia}}}

\affil*[3]{\orgdiv{School of Computing and Creative Technologies}, \orgname{University of the West of England}, \orgaddress{\city{Bristol}, \country{UK}}}


\abstract{

The cognitive and reasoning abilities of large language models (LLMs) have enabled remarkable progress in natural language processing. However, their performance in interpreting structured data, especially in tabular formats, remains limited. Although benchmarks for English tabular data are widely available, Arabic is still underrepresented because of the limited availability of public resources and its unique language features. To address this gap, we present AraTable, a novel and comprehensive benchmark designed to evaluate the reasoning and understanding capabilities of LLMs when applied to Arabic tabular data. AraTable consists of various evaluation tasks, such as direct question answering, fact verification, and complex reasoning, involving a wide range of Arabic tabular sources. Our methodology follows a hybrid pipeline, where initial content is generated by LLMs and subsequently filtered and verified by human experts to ensure high dataset quality. Initial analyses using AraTable show that, while LLMs perform adequately on simpler tabular tasks such as direct question answering, they continue to face significant cognitive challenges when tasks require deeper reasoning and fact verification. This indicates that there are substantial opportunities for future work to improve performance on complex tabular reasoning tasks. We also propose a fully automated evaluation framework that uses a self-deliberation mechanism and achieves performance nearly identical to that of human judges. This research provides a valuable, publicly available resource and evaluation framework that can help accelerate the development of foundational models for processing and analysing Arabic structured data.}
\keywords{Arabic Benchmark, Tabular Data, Automatic Evaluation, LLMs' Cognitive Capacity}



\maketitle

\section{Introduction}

Large language models (LLMs) have demonstrated remarkable capabilities across a wide range of natural language processing (NLP) tasks, achieving state-of-the-art performance in areas such as text generation \cite{li2024dtllm,liu2024shield}, translation \cite{koshkin2024transLlama,wang2024benchmarking}, and question answering \cite{allemang2024increasing,bui2024cross}. Their ability to process and understand complex textual information has opened up new avenues for the automation and enhancement of various data-driven applications. However, the performance of LLMs is heavily reliant on the nature and structure of the input data. While significant progress has been made in handling unstructured text, evaluating and improving LLM performance on structured data, particularly tabular data, remains a critical area of research.

Tabular data is widely used in various domains, including finance, healthcare, and public administration, and contains valuable information, organised in rows and columns \cite{o2024literature}. Extracting insights and answering questions from tabular data requires models to understand not only the natural language query but also the table structure, the relationships between columns and rows, and the specific values within individual cells. Evaluating LLMs' performance on tasks involving tabular data, such as question answering based on the information presented in tables, fact verification against tables, and complex reasoning based on tabular information, presents unique challenges \cite{Nguyen2024InterpretableLT,Zhang2024NormTabIS,Wang2025TableLE}. These tasks require capabilities beyond simple text comprehension, including symbolic reasoning, data aggregation, comparison, mathematical calculation, and logical deduction based on structured facts.

Although there are general benchmarks for evaluating LLMs' performance on various tasks, with some recent focus on tabular data in English, such as TableBench \cite{Wu2024TableBenchAC}, DataBench \cite{OsesGrijalba2024QuestionAO}, and more recently, MMTU \cite{xing2025mmtu}, there is a significant gap in comprehensive benchmarks specifically designed to evaluate LLMs' performance on Arabic tabular data. The Arabic language presents its own set of linguistic complexities, including rich morphology, various dialects, and complex grammar, which can impact an LLM's ability to process and understand Arabic text, and, consequently, Arabic tabular data. Existing Arabic LLM evaluation efforts, such as the Open Arabic LLM Leaderboard\footnote{\url{https://huggingface.co/OALL}}, often focus on general language understanding, with limited attention paid to the challenges of handling structured Arabic information.

To address this critical gap, this study introduces \textbf{AraTable}, a novel benchmark specifically designed to evaluate the performance of LLMs on Arabic tabular data. This benchmark focuses on three key tasks: direct question answering, fact verification, and reasoning, covering a range of complexity levels and requiring different types of interaction with the tabular structure and content. We considered diverse datasets of Arabic tables from various sources, including Wikipedia, real-world data, and LLM-generated data. Through a comprehensive analysis using AraTable, we evaluated the performance of several prominent LLMs, specifically Llama 3.3 70B \cite{grattafiori2024Llama}, Mistral Large\cite{mistral2024large}, DeepSeek-V3\cite{liu2024deepseek}, and Jais 70B \cite{sengupta2023jais}. Across all datasets, our evaluation shows that DeepSeek-V3 consistently outperformed other models, closely followed by Llama 3.3 70B and Mistral Large, while Jais models showed significantly lower accuracy. Direct question-answer (QA) tasks were consistently easier for all models, highlighting a clear gap in performance for more complex reasoning questions, where accuracy often remained below 60\%. This indicates a fundamental limitation in current LLMs' ability to perform complex logical inferences over Arabic tabular data, and that the pure Arabic models, as seen with Jais, are insufficient for complex tabular reasoning.

Additionally, we introduce and validate a robust evaluation framework specifically designed to assess LLMs' free text answers and compare them with ground-truth answers. As illustrated in Figure \ref{fig:Evaluation}, we first considered a multi-round human evaluation to ensure robustness. Then, we proposed an assisted self-deliberation (ASD) mechanism for LLM-based automatic evaluation. The goal is to provide a reliable, unbiased, automatic evaluation process of the models' answers for AraTable benchmarking that closely aligns with human judgement.

This work provides a valuable resource for the research community, enabling more targeted development and evaluation of LLMs for Arabic tabular data processing and contributing to the advancement of Arabic natural language understanding in the structured domains. The main contributions of this study can be summarised as follows:

\begin{itemize}

  \item Construct a benchmark dataset of Arabic tabular data specifically designed to assess LLM cognitive capabilities in direct question answering, fact verification, and reasoning.

    \item Develop a hybrid benchmark generation and validation process through which LLMs produce the initial content, which is rigorously filtered and verified by native Arabic-speaking experts to ensure high quality and facilitate evaluation of LLM performance.

    \item Incorporate Arabic tabular data from diverse sources into one dataset to assess LLM robustness across different domains and sources of Arabic tables.

    \item Propose ASD, a robust automatic evaluation method that adopts self-deliberation mechanisms that align closely with human judgement to measure LLMs' performance on AraTable accurately.

    \item Analyse and compare the performance of various LLMs on the developed benchmark dataset, identifying their strengths and weaknesses in handling these complex tasks.

    \item Provide a publicly available dataset promoting further research and development in Arabic tabular data processing.

\end{itemize}

The remainder of this paper is organised as follows: Section \ref{sec:RW} describes the main related works aligned with this study. Section \ref{Sec:Benchmark} presents the construction details of AraTable. We describe the experimental setting to assess the utility of AraTable in Section \ref{sec:Experement}, we introduce our comprehensive evaluation method in Section \ref{sec:Eval}, and we thoroughly analyse and discuss the results in Section \ref{sec:resulta}. Finally, we present the main conclusions of our study in Section \ref{sec:Concl} and outline some of the study's limitations in Section \ref{sec:limit}.

\section{Related Work}\label{sec:RW}

Given the novelty of the QA task on Arabic tabular data with LLMs, in this section, we will review the main related works, first presenting studies relating to the evaluation of LLMs on Arabic language tasks, then studies focused on LLMs' understanding of tabular data. Finally, we will discuss studies that have deployed LLMs as automatic evaluators.

\subsection{Evaluation of LLMs in Arabic}

The evaluation of LLMs in Arabic is a rapidly evolving yet complex domain, driven by Arabic's significant global presence and its unique linguistic complexities, including rich morphology, diverse dialects, and intricate syntactic structures \cite{Rhel2025}, \cite{Mousi2025}. Although significant progress has been made in the development of Arabic-specific LLMs, such as Jais \citep{sengupta2023jais}, a comprehensive evaluation remains a challenge. Existing benchmarks, such as AraBench \citep{Sajjad2020}, ALUE \citep{seelawi-etal-2021-alue}, ARLUE \citep{AbdulMageed2021}, and, more recently, AraReasoner \cite{hasanaath2025arareasoner}, assess general NLP tasks such as sentiment analysis and summarisation. \citet{abdallah2024arabicaqa} introduced ArabicaQA, a comprehensive dataset designed to evaluate LLM performance in machine reading comprehension and open-domain question answering. Other key evaluation aspects specific to Arabic remain underrepresented. For instance, recent studies have addressed hallucination and faithfulness issues \cite{abdaljalil2025halluverse25}. Similarly, fairness and bias have been evaluated by \citet{alghamdi2025aratrust}. Furthermore, comprehensive initiatives such as BALSAM \footnote{\url{https://benchmarks.ksaa.gov.sa/b/balsam/tasks}} and the aiXplain Arabic LLM Benchmark Report \footnote{\url{https://aixplain.com/arabic-llm-benchmark-report/}} aim to standardise multitask evaluations, emphasising cultural understanding. However, to the best of our knowledge, none of the previous works included structured data inputs or evaluated the LLMs' ability to understand the structured data. Given that our work introduces an Arabic LLM evaluation benchmark, named AraTable, tailored specifically to tabular data, it aims to complement these prior efforts by addressing the notable gap concerning structured-input evaluation.

\subsection{LLMs and Tabular Data Understanding}

The ability of LLMs to interpret and reason using structured tabular data, which presents unique challenges compared to unstructured text, has attracted increasing research attention. \citet{sui2024table} introduced a benchmark to evaluate LLMs' structural understanding of tables, including tasks such as cell lookup and row retrieval. They show that input formatting significantly affects performance and propose a self-augmented prompting strategy to improve it. Similarly, \citet{liu-etal-2024-rethinking} show that LLMs struggle with structural variations in tables, even when the content remains unchanged. To address such limitations, recent efforts have moved beyond generic prompting. For instance, \citet{ziqi2023tab} introduce Tab-CoT, a prompting strategy that organises intermediate reasoning steps in a table format to guide inference over unstructured text. Building upon this, \citet{wang2024chain} propose Chain-of-Table, a method that guides LLM reasoning by iteratively applying structured operations, such as filtering and aggregation, to progressively transform the table. Instead of directly generating the final answer, the model constructs a sequence of intermediate table modifications aimed at deriving the correct answer. Representation modality is another important factor that influences LLM performance on tabular tasks. \citet{deng2024tables} compare text-linearised, JSON, and image-based table inputs, revealing performance variations across formats.

In terms of evaluating LLMs on question answering from tabular data, \citet{chen-2023-large} investigated the capabilities of LLMs for table reasoning using few-shot in-context learning. Their study shows that LLMs, when combined with Chain-of-Thought prompting, perform well on table QA and fact verification tasks. In a related effort, \citet{grijalba2024question} introduces DataBench, a benchmark covering 65 real-world datasets across diverse domains, to support large-scale evaluation. Their results indicate that, despite recent progress, both open- and closed-source models still face challenges, even with relatively simple questions. \citet{bhandari2024robustness} further analyses the robustness of LLMs on table QA by studying the effects of in-context learning, model scale, instruction tuning, and domain biases across multiple datasets. They find that instruction tuning improves performance, but data contamination and domain-specific complexity continue to hinder reliability. Recently, \citet{xing2025mmtu} introduced MMTU, a new benchmark comprising over 30,000 questions across 25 real-world table tasks. MMTU is designed to thoroughly assess a model’s ability to understand, reason with, and manipulate tables at an expert level. Preliminary evaluations using MMTU show that even advanced LLMs such as OpenAI’s GPT-4o-mini and DeepSeek R1 achieve only around 60\% accuracy, highlighting significant room for improvement in handling QA over structured data.

Despite the significant advancements highlighted in previous research, a consistent finding across these studies is that LLM performance on structured tabular data is still far from optimal. Furthermore, nearly all existing evaluations have concentrated solely on the English language. To our knowledge, no prior work has yet assessed the performance of LLMs on tabular question answering and fact verification in Arabic, revealing a crucial research gap that this work directly aims to address.

\subsection{LLMs as Automatic Evaluators}

The increasing complexity and subjectivity of language generation tasks have led to increased interest in using LLMs as scalable and cost-effective automatic evaluators, sometimes referred to as LLMs-as-Judges \cite{li2024llms}. LLM-based evaluation systems have been categorised by \citet{li2024llms} into three main setups: (1) single-LLM systems that rely on only one LLM evaluator, as in the work carried out by \citep{lin2023llm} and \citep{liu2023g}; (2) multi-LLM systems that combine multiple LLMs to perform the evaluation task \citep{chan2023chateval}, \citep{chu2024pre}; and (3) hybrid systems that combine LLMs with human evaluators \citep{li2023collaborative} \citep{ma2025towards}.

Among others, \citet{lee2025evaluating} investigated how consistently LLMs produce evaluation scores, revealing notable variation across different prompts and sampling settings. Following this direction, \citet{panickssery2024llm} examined self-preference in LLM evaluators, which is the tendency to favour their own outputs over others. The authors identify a correlation between self-recognition and self-preference. Fine-tuning experiments suggest a causal relationship, raising concerns about the fairness of self-evaluation. \citet{zhang2024large} explored the use of LLMs to assess the quality of recommendation explanation texts based on real user feedback. They propose a three-level meta-evaluation strategy and demonstrate that zero-shot LLMs can match or outperform traditional metrics such as BLEU and ROUGE. In addition, they show that the combination of multiple LLMs or the integration of human feedback improves the accuracy and stability of the evaluation. 

In this work, we present a novel LLM-based automatic evaluation framework that achieves strong alignment with human judgements on the AraTable benchmark. Our approach, which we call assisted self-deliberation (ASD), is based on a self-deliberation philosophy in which two independent LLMs evaluate the correctness of the answers. However, instead of requiring full answer regeneration, ASD activates only in cases of disagreement, prompting each model to revisit its own decision while also considering why the other model might have reached a different conclusion. Crucially, this process relies solely on a disagreement signal, without disclosing the reasoning of the opposing model or engaging in the multi-turn debates common in existing work. This novel, lightweight strategy requires just one re-evaluation pass per judge in cases of disagreement, with each model justifying both perspectives using the shared rubric.

\section{AraTable: Benchmarking LLMs on Arabic Tabular Data}\label{Sec:Benchmark}

\begin{figure}[h]
\centering
\includegraphics[clip=true,scale=.95]{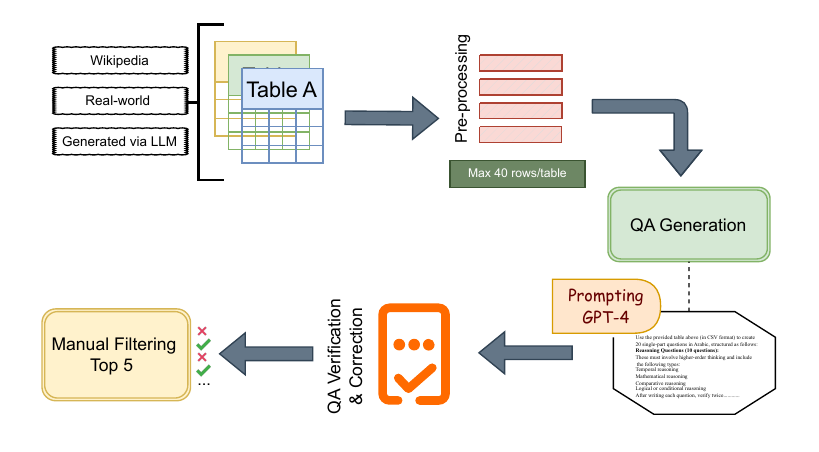}
\caption{Overview of AraTable construction process per task}
\label{fig:Illustration2}
\end{figure}

In this section, we introduce AraTable, a benchmark designed to evaluate the capabilities of LLMs on QA tasks over Arabic tabular data. Figure~\ref{fig:Illustration2} provides an overview of the benchmark construction pipeline. 

AraTable is constructed with a focus on supporting diverse reasoning types, including arithmetic, logical, and comparative inference over Arabic tables. In addition to reasoning-based questions, AraTable includes direct answer extraction and table-based fact verification tasks. Incorporating all three enables a comprehensive assessment of model capabilities, ranging from surface-level understanding to deeper analytical reasoning and verification. This multitask structure aligns with prior efforts such as WikiTableQuestions~\cite{pasupat2015compositional}, TAT-QA~\cite{zhu2021tat}, and TabFact~\cite{chen2019tabfact}, which collectively demonstrate the value of combining different question types to evaluate reasoning depth and generalisation. The following subsections present a description of data sourcing, QA generation, and manual validation.

\begin{table}[h]
    \renewcommand{\arraystretch}{1.2} 
    \centering
    \begin{tabular}{ |p{0.4cm}||p{8cm}|p{1cm}|p{1.2cm}|}
     \hline
     & Table Name & Rows & Columns\\
     \hline
     1 & \vspace*{-2.5mm}\RL{قائمة أكثر متاحف الفنون زيارة في العالم} & 100 & 5\\
     2 & \vspace*{-2.5mm}\RL{قائمة الأطباق العربية} & 84 & 4\\
     3 & \vspace*{-2.5mm}\RL{قائمة الأندية أبطال العالم} & 67 & 9\\
     4 & \vspace*{-2.5mm}\RL{قائمة الحاصلين على جائزة نوبل} & 125 & 7\\
     5 & \vspace*{-2.5mm}\RL{قائمة الدول حسب تعداد وسائل النقل لكل 1000 شخص} & 192 & 4 \\
     6 & \vspace*{-2.5mm}\RL{قائمة السيارات الأكثر مبيعا} & 69 & 4 \\
     7 & \vspace*{-2.5mm}\RL{قائمة الشركات حسب الدخل} & 188 & 9 \\
     8 & \vspace*{-2.5mm}\RL{قائمة اللاعبين الفائزين بكأس العالم} & 422 & 7 \\
     9 & \vspace*{-2.5mm}\RL{قائمة المدن المستضيفة للألعاب الأولمبية} & 57 & 10 \\
     10 & \vspace*{-2.5mm}\RL{قائمة مساجد حول العالم} & 163 & 5 \\
     11 & \vspace*{-2.5mm}\RL{قائمة مهرجان دبي لأهم مئة فيلم عربي} & 100 & 5 \\
     12 & \vspace*{-2.5mm}\RL{قائمة مواقع التراث العالمي في الدول العربية} & 112 & 6 \\
     13 & \vspace*{-2.5mm}\RL{قائمة أقدم البنايات في العالم} & 92 & 6 \\
     14 & \vspace*{-2.5mm}\RL{قائمة الدول} & 207 & 4 \\
     15 & \vspace*{-2.5mm}\RL{قائمة رؤساء الولايات المتحدة} & 46 & 6 \\
     16 & \vspace*{-2.5mm}\RL{قائمة أطول المباني في العالم} & 43 & 8 \\
     17 & \vspace*{-2.5mm}\RL{التصنيف الأكاديمي لجامعات العالم} & 121 & 13 \\
     18 & \vspace*{-2.5mm}\RL{قائمة أكبر الجامعات في العالم من حيث عدد الملتحقين بها} & 88 & 9 \\
     19 & \vspace*{-2.5mm}\RL{أكبر شركات الطيران في العالم} & 10 & 8 \\
     20 & \vspace*{-2.5mm}\RL{ترتيب ويبوميتركس لجامعات العالم} & 25 & 3 \\
     21 & \vspace*{-2.5mm}\RL{تشكيلة المنتخبات الفائزة بكأس العالم} & 22 & 5 \\
     22 & \vspace*{-2.5mm}\RL{قائمة أكبر الخسائر للشركات} & 18 & 11 \\
     23 & \vspace*{-2.5mm}\RL{قائمة أكبر الأرباح للشركات} & 34 & 8 \\
     24 & \vspace*{-2.5mm}\RL{قائمة الدول العربية حسب عدد السكان} & 22 & 10 \\
     25 & \vspace*{-2.5mm}\RL{قائمة أسرع السيارات بنسبة التسارع} & 62 & 5 \\
     26 & \vspace*{-2.5mm}\RL{قائمة أكبر مطارات العالم حسب حركة الركاب} & 50 & 8 \\
          
    \hline
    \end{tabular} 
    \caption{List of Wikipedia Arabic tables that are considered in this research }
    \label{tab:Wikitable}
\end{table}

\subsection{Tabular Data Collection}\label{Sec:TabData Collection}

The benchmark aims to enable comprehensive evaluation across a wide range of scenarios by considering tables from diverse domains, such as tourism, transportation, sport, business, politics, education, hospitality, demographics, entertainment, and government. This diversity ensures broad domain coverage, which facilitates the assessment of different reasoning capabilities within the Arabic language context and follows recent benchmark design practices that emphasise coverage across multiple domains to support comprehensive model evaluation~\cite{wu2025tablebench}.

The tables were sourced from a range of platforms, including:
\begin{itemize}

  \item \textbf{Wikipedia:} Tables extracted from Arabic Wikipedia pages provide a foundational source of structured knowledge across various domains and are widely used in table QA benchmarks due to their consistent formats and topic diversity~\cite{pasupat2015compositional,kweon2023open}. A complete list of Wikipedia tables that we considered in this study, along with their dimensions (numbers of rows and columns), is presented in Table~\ref{tab:Wikitable}.

    \item \textbf{Real-World Data:} Tables were obtained from platforms such as Kaggle, national open data portals, and government agencies. These cover various topics, including hospitality, real estate, automotive, literature, weather, and public events. A complete list is provided in Table~\ref{tab:Real-worldtable}. Incorporating real-world tables complements Wikipedia content by introducing naturally occurring structures, complex formatting, and domain-specific variability. This is in line with recent benchmarks such as TAT-QA~\cite{zhu2021tat} and TableEval~\cite{zhu2025tableeval}, which incorporate tables from real-world sources.

    \item \textbf{LLM-Generated Data:} To ensure coverage of scenarios not adequately represented in available real-world sources, additional tables were generated using GPT-4o based on structured prompts. Recent research suggests that augmenting benchmarks with LLM-generated tables can effectively address domain gaps and cover rare or edge-case scenarios~\cite{berkovitch2024generating}. These supplementary tables enhance the comprehensiveness of the benchmark. Table~\ref{tab:LLMstable} shows the list of tables.

\end{itemize}

\begin{table}[h]
    \begin{tabular}{|p{0.4cm}||p{6cm}|p{1cm}|p{1.1cm}||p{2.9cm}|}
     \hline
     & Table Name & Rows & Columns & Source\\
     \hline
     1 & HARD: Hotel Arabic-Reviews Dataset & 93700 & 7 & \citeauthor{elnagar2018hotel}\footnotemark[1] \\
     2 & Saudi Arabia Used Cars Dataset & 8248 & 14 & Kaggle\footnotemark[2]\\
     3 & Jamalon Arabic Books Dataset & 8986 & 11 & Kaggle\footnotemark[3]\\
     4 & Saudi Arabia Real Estate (AQAR) & 3718 & 24 & Kaggle \footnotemark[4]\\
     5 & \vspace*{-2.5mm}\RL{رحلات القطار}  & 26 & 8 & SAR \footnotemark[5]\\
     6 & \vspace*{-2.5mm}\RL{اطوال الطرق حسب التصنيف الهندسي} & 14 & 6 & Open Data Platform \footnotemark[6]\\
     7 & Events 2024   & 2637 & 10 & Open Data Platform \footnotemark[6] \\
     8 & \vspace*{-2.5mm}\RL{المناقصات}  & 45 & 7 & Open Data Platform \footnotemark[6] \\
     9 &  NUMBER OF THUNDERSTORMS OBSERVED BY PME MET STATIONS-2006  & 29 & 13 & National Center for Meteorology \footnotemark[7] \\
     10 & \vspace*{-2.5mm}\RL{النسبة المئوية لمجالات الكتب لدى الاسر التي لديها مكتبة منزلية حسب المنطقة الإدارية}  &  14 & 14 &General Authority for Statistics \footnotemark[8] \\
     \hline
    \end{tabular}
    \caption{List of Real-world Arabic tables considered in this research}
    \label{tab:Real-worldtable}

\end{table}
\interfootnotelinepenalty=10000 

\footnotetext[1]{\url{https://github.com/elnagara/HARD-Arabic-Dataset}}
\footnotetext[2]{\url{https://www.kaggle.com/datasets/turkibintalib/saudi-arabia-used-cars-dataset}}
\footnotetext[3]{\url{https://www.kaggle.com/datasets/dareenalharthi/jamalon-arabic-books-dataset}}
\footnotetext[4]{\url{https://www.kaggle.com/datasets/lama122/saudi-arabia-real-estate-aqar}}
\footnotetext[5]{\url{https://www.sar.com.sa/}}
\footnotetext[6]{\url{https://data.gov.sa/}}
\footnotetext[7]{\url{https://ncm.gov.sa/Ar/MediaCenter/OpenData/}}
\footnotetext[7]{\url{https://www.stats.gov.sa/statistics-tabs?tab=436312&category=1340120}}

\begin{table}[h!]
    \renewcommand{\arraystretch}{1.2} 
    \centering
    \begin{tabular}{ |p{0.4cm}||p{7cm}|p{1cm}|p{1.2cm}|}
     \hline
     & Table Name & Rows & Columns\\
     \hline
     1 & \vspace*{-2.5mm}\RL{أداء الطلاب} & 20 & 9 \\
     2 & \vspace*{-2.5mm}\RL{إدارة الرواتب والدوام} & 25 & 9 \\
     3 & \vspace*{-2.5mm}\RL{مصاريف شهرية} & 30 & 6 \\
     4 & \vspace*{-2.5mm}\RL{جدول إجازات سنوية} & 30 & 7 \\
     5 & \vspace*{-2.5mm}\RL{تكلفة المنتجات} & 30 & 10 \\   
    \hline
    \end{tabular} 
    \caption{List of LLMs generated Arabic tables that are considered in this research}
    \label{tab:LLMstable}
\end{table}

\subsection{Table Pre-processing}

Given the diversity of table sources, all tables were standardised into a unified format to ensure consistency. Original table names were preserved. Tables not originally in CSV format, or those with incompatible structures, were manually adjusted. Each table was also limited to a maximum of 40 rows to accommodate model input size constraints, specifically addressing the limitations of the Jais model, while shorter tables remained unchanged. This is a common consideration when adapting tabular data for LLMs~\cite{fang2024large}.

\subsection{QA Generation }

We employed a prompt-based approach using GPT-4 to generate an initial batch of QA pairs for each table in the dataset. The QA generation process consisted of three distinct tasks: direct question answering, reasoning-based questions, and fact verification. For each task, GPT-4 produced 10 questions per table, resulting in a total of 30 questions per table. Each task is described in the following:

\begin{table}[h!]
\centering
\begin{tabular}{|p{5cm}|p{1cm}|p{2.14cm}|p{2.57cm}|}
\hline
\textbf{Question} & \textbf{Answer} &  \textbf{Task} & \textbf{Reasoning Type}   \\
\hline
\RL{أي مدينة استضافت الألعاب الأولمبية الصيفية لعام 1960؟} & \RL{روما} & Simple QA & -  \\
\hline
 \RL{ما هي المدينة التي استضافت الألعاب الأولمبية الصيفية بعد أول دورة شتوية؟}& \RL{ باريس}& Reasoning QA & Temporal reasoning \\
\hline
\RL{استضافت اليابان الألعاب الأولمبية الشتوية مرتين قبل عام 1980} &\RL{خطأ} & Fact Verification & - \\
\hline
\end{tabular}
\caption{Examples from AraTable illustrating the three QA tasks: simple, reasoning, and fact verification based on a table sourced from Wikipedia.}
\label{tab:benchmark-examples}
\end{table}

\vspace{1em}
\noindent\textbf{Direct QA}
This task focuses on the direct retrieval of factual information from tables without requiring inference. The questions are designed to evaluate whether a model can accurately locate and extract specific values or entries from the table content.

\vspace{1em}
\noindent\textbf{Reasoning QA}
Reasoning questions require a deeper understanding of the tabular data through multiple inference steps. They cover various types of reasoning: temporal reasoning, which involves interpreting and comparing time-related data such as dates or durations; mathematical reasoning, which requires performing calculations such as totals and differences; comparative reasoning, which focuses on comparing entries; and logical or conditional reasoning, which involves evaluating conditions or constraints across multiple fields. Table \ref{tab:reasoning_counts} shows the statistics of each reasoning type occurring in the dataset. The prompt used to generate both direct and reasoning QA is provided in Prompt~\ref{prompt:simpleandreasoning}.
\begin{promptbox}[label={prompt:simpleandreasoning}]{Simple and Reasoning QA}\small
Use the table provided above (in CSV format) to create 20 single-part questions in Arabic, structured as follows:\\
\textbf{Reasoning Questions (10 questions):} These must involve higher-order thinking and include the following types:\\
* Temporal reasoning\\
* Mathematical reasoning\\
* Comparative reasoning\\
* Logical or conditional reasoning\\
After writing each question, verify twice that it accurately reflects the intended reasoning type. If it does not, revise the question accordingly.\\

\textbf{Non-reasoning Questions (10 questions):} These should be straightforward, answerable directly from the table without inference or calculation.\\
Total Number of Questions: 20\\

\textbf{Output Format:} Save the output as a CSV file (UTF-8 with BOM encoding) with the following columns:\\
* Question: Written in Modern Standard Arabic\\
* Type of Question: Either "Reasoning" or "Non-Reasoning"\\
* Type of Reasoning: Leave this column blank for non-reasoning questions. For reasoning questions, specify the reasoning type (e.g., "Mathematical reasoning")\\
* Answer: Provide a short, specific answer derived directly from the table (use numbers only, without extra text, if applicable)\\
* Description: A brief explanation of how the answer was extracted from the table\\

\textbf{Additional Notes:}
* All questions must be strictly based on information explicitly present in the table (in Arabic). Do not introduce any assumptions, inferred facts, or external knowledge.\\
* Begin with the 10 reasoning questions, followed by the 10 non-reasoning questions.\\
* Ensure effective utilisation of the table's structure for each question.

\end{promptbox}

\vspace{1em}
\noindent\textbf{Fact Verification}
The fact verification task assesses whether a given statement is true or false based on the content of the table. The prompt used to generate fact verification QA pairs is provided in Prompt~\ref{prompt:Verification}.

\vspace{1em}
\noindent Table~\ref{tab:benchmark-examples} shows an example QA pair for each task from the Olympic Host Cities table, sourced from Arabic Wikipedia. The table contains information about the city, country, continent, year, start and end dates, season, and edition of each Summer Olympic Games. Each example is labelled with its corresponding task and, where applicable, the reasoning type.

\begin{promptbox}[label={prompt:Verification}]{ Fact Verification}\small

Use the table provided above (in CSV format) to create 10 truly complex True/False questions that require deep reasoning, such as:\\
* Combining multiple conditions\\
* Cross-referencing multiple columns\\
* Logical deduction based on the provided data\\
Total Number of Questions: 10 (five with the answer "True" and five with the answer "False")\\

\textbf{Output Format:} Save the output in a CSV file (UTF-8 with BOM encoding) with the following columns:\\
* Question: A statement written in Modern Standard Arabic.\\
* Answer: "True" or "False".\\
* Description: A concise explanation of how the answer was derived from the table\\

\textbf{Additional Notes:}
* Each statement must be based strictly on information explicitly present in the table (in Arabic). Do not introduce any assumptions, inferred facts, or external context.\\
* Ensure that the structure of the table is effectively utilised in all questions.

\end{promptbox}

\subsection{ Manual Filtering and Verification}

Following generation, all questions and answers were manually reviewed for clarity, correctness, and relevance. Each answer was independently verified by three human annotators using a combination of code, Excel functionality, and manual inspection to ensure consistency and factual alignment with the corresponding table. Both questions and answers were corrected if necessary. For each task, only five questions were retained, selected to target information across different columns and rows. This filtering step ensured that the final set was accurate and representative. As a result, each table included a total of 15 QA pairs.

\subsection{Dataset Statistics }

The final benchmark comprised 41 tables, each paired with 15 QA pairs: five for each of the direct QA, reasoning QA, and fact verification tasks, totalling 615 human-validated QA pairs. The number of questions per reasoning type across the dataset is presented in Table~\ref{tab:reasoning_counts}. Our dataset comprised 205 reasoning questions, with mathematical reasoning accounting for the majority, at 79 questions, nearly 39\% of the total. Comparative reasoning comprised 67 questions, logical reasoning had 42 questions, and temporal reasoning, the smallest type, accounted for just over 8\%, with 17 questions. This uniform task distribution allowed for fair comparisons across question types. The benchmark is publicly available for research purposes at \footnote{\url{https://github.com/rana-alshaikh/AraTable-Benchmark}}.

\begin{table}[ht!]
\centering
\renewcommand{\arraystretch}{1.2}
\begin{tabular}{|l|c|}
\toprule
Reasoning Type & Number of Questions \\
\midrule
Mathematical Reasoning & 79 \\
Comparative Reasoning & 67 \\
Logical Reasoning & 42 \\
Temporal Reasoning & 17 \\
\bottomrule
\end{tabular}

\caption{Number of Questions per Reasoning Type }
\label{tab:reasoning_counts}
\end{table}

\section{Experimental Setting}\label{sec:Experement}
 We employed the following experimental setup to evaluate LLM performance on our proposed benchmark. 

\subsection{Comparison models} Our focus was on Arabic-aware open-source models. We chose four models with strong performance in Arabic tasks: Llama 3.3 70B \cite{grattafiori2024Llama}, Mistral Large\cite{mistral2024large}, DeepSeek-V3\cite{liu2024deepseek}, and Jais 70B \cite{sengupta2023jais}.

\subsection{Zero-Shot In-Context Learning}

We gave each LLM three sequential parts, in line with the \citet{grijalba2024question} method: The task description comprised directions that defined the objective: "Answer the following questions using only the information in the table below and return the answer without any explanation".

Second, tabular data was provided as a CSV with headers and original cell values. By giving the table in CSV format, we ensured that the models processed it as structured text without requiring additional preprocessing. The third part was the question set, a series of questions that referenced the table contents. Each question was written in such a way that the answer could be found using one or more cells in the table. Formatting the prompt in this way was an attempt to restrict the models to using the table as the sole source of information from which to answer the questions, thereby minimising contamination from external knowledge within the model. This enabled us to assess the model's ability to understand tables written in Arabic in a reproducible manner and under consistent, controlled conditions.

In our setting, no constraints were imposed on the format of the answer; models were free to respond in plain text, tables, lists, or any other structure; therefore, no type suffixes were added to any prompts except in the fact-verification statements, which were considered to be Boolean questions. For these, we prompted the model to answer \texttt{"True"} or \texttt{"False"}. There were several reasons behind the choice not to impose any restrictions. First, it helps to decrease prompt bias: by not requiring a specific output structure, we avoided enforcing models towards a particular representation, which allowed us to capture their natural preferences for organising and presenting the output\cite{nguyen2025llms}. In support of this, \cite{tam2024let} finds that, in reasoning tasks, generally, we can attain better performance by relaxing the format restrictions.

Additionally, as our research focused only on open-source models, it is crucial to underline that these models typically comply less with strict output formats in comparison with closed-source models, as \cite{xia2024fofo}, highlighting that imposing format restrictions on open-source models usually fails to produce the intended advantages, so supporting our option. 

Lastly, this approach reflects practical deployment scenarios in which users usually want human-readable responses instead of exact machine-readable outputs. By allowing the models to choose their preferred response format, we gained insight into their ability to provide accurate and effective responses in a flexible and realistic setting \cite{lakkaraju2022rethinking}.
These factors ensured that our evaluation focused solely on the models' ability to comprehend Arabic tabular data, rather than their behavioural compliance with arbitrary formatting constraints.
Even though the models were instructed to respond briefly, Jais' responses were often verbose, filled with unnecessary repetition and explanations. Consequently, for Jais, we used two collection methods: (1) saving the original, unmodified outputs and (2) having two independent human annotators extract the concise answer from the lengthy response. Model outputs were saved in CSV format for later evaluation.

\section{Evaluation}\label{sec:Eval}

Allowing the model to answer freely without following any structure imposed challenges in the evaluation phase as these models were significantly different in terms of verbosity and response form: some of the models gave very brief, concise answers, as instructed, while others provided excessively detailed explanations using various synonyms for the words in the questions. For example, instead of answering with True, the model used phrases such as  "Indeed \RL{بالفعل}" or "This is a fact \RL{هذه حقيقة}."

Additionally, we observed that some of these models were more sensitive to minor alterations in the listing, numbering, and grouping of the questions. For example, we found that Jais and Mistral could give unordered answers or even miss one of the questions if the questions were not numbered. Moreover, Jais occasionally filtered responses based on trigger words or transitions to English mid-answer, especially in entity names, which could indicate that the model does not depend solely on the information provided in the table to answer the question. Another challenge was that the models frequently exhibited inconsistent normalisation of punctuation and diacritics, as well as text alignment.\\

\begin{figure}
\centering
\includegraphics[width=\linewidth]{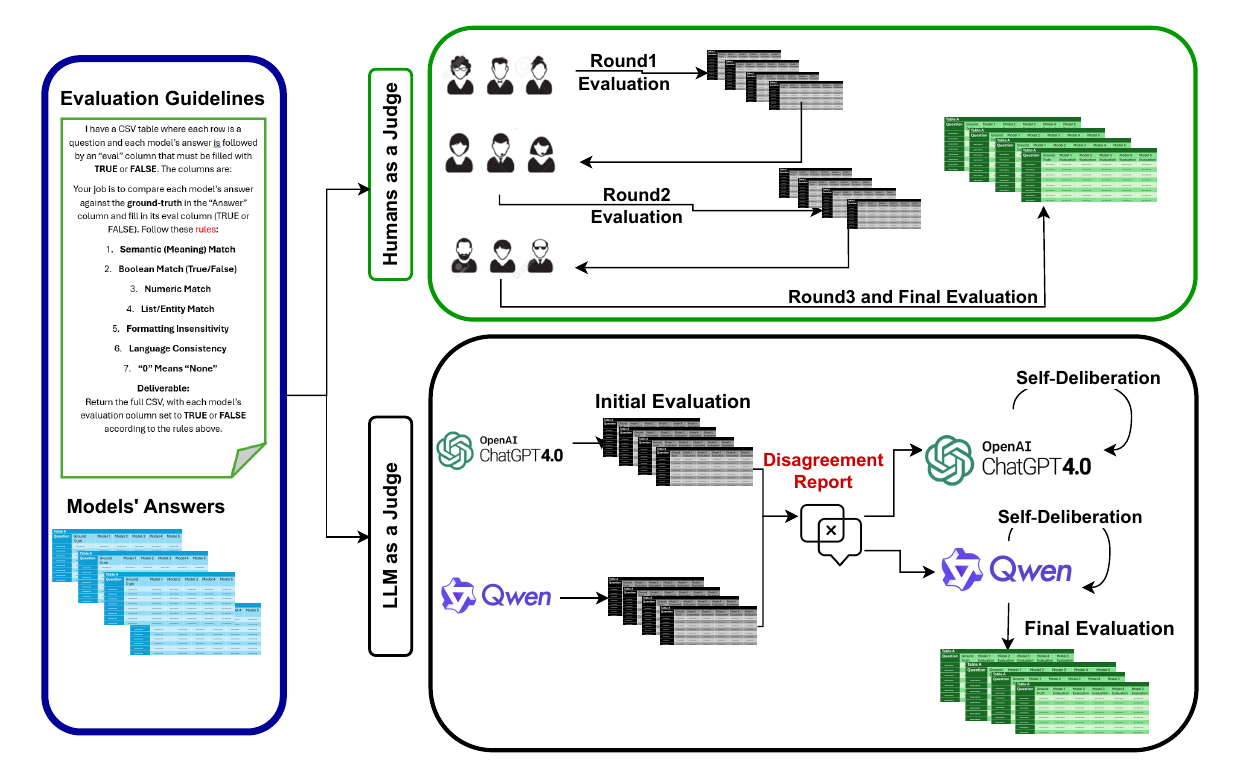}
\caption{Illustration of the comprehensive evaluation framework for AraTable, using human judges and LLM judges.}
\label{fig:Evaluation}
\end{figure}

In light of these observations, the unconstrained settings and the resulting challenges and inconsistencies suggest that employing automated scoring methods, such as exact matching or embedding similarity, may not be ideal for our setting. Even in situations in which we can impose stringent constraints on the model answer, these methods partially test the degree to which the models adhere to the given instructions rather than being entirely focused on the models' understanding of the tabular data. Consequently, in our evaluation framework, we imposed a relaxed evaluation, allowing us to focus on measuring the models' ability to understand Arabic tabular data.
\\
\\
We designed an evaluation framework to assess the responses of the four models, which were, in total, five distinct outputs, as Jais had two versions of the responses: a summarised answer and a full-length response. During the evaluation phase, all models were anonymised by assigning numeric identifiers (e.g., Model 1, Model 2). Furthermore, the full-length Jais outputs and their human-extracted answers were treated as separate entities with distinct identifiers, assigned as Model 4 and Model 5, respectively. This design enabled (a) evaluation of the accuracy of the human-extracted response from Jais' verbose output and comparison of human evaluation consistency when presented with the same answer in both concise (Model 4) and verbose (Model 5) forms – ideally, both should yield the same accuracy, as the core answer remains unchanged; and (b) the assessment of the robustness of our automated evaluation method in understanding and evaluating the LLMs' responses that exhibited verbosity. Figure ~\ref{fig:Evaluation} provides an overview of the evaluation framework pipeline.
\subsection{Human as Judges}\label{subsec:HumanasJudge}
To establish a reliable baseline, nine native Arabic-speaking human evaluators were assigned the role of judges. The human evaluators were organised into three groups to evaluate the models' answers over three consecutive rounds, using a detailed rubric to determine correctness. The first group independently evaluated every LLM-generated answer against the ground-truth annotations. The second group then reviewed the evaluation, and in round 3, the third group reviewed the evaluation and generated the final assessment. 

In the used rubric, one of the rules was that if the answer conveyed the same meaning as the ground-truth, then it was considered correct, i.e., an exact string match with the ground-truth was not required. This guideline allowed for any variations in wording or the addition of descriptive words or units. For example, \textbf{"around 40," "40," and "forty years"} were all considered equivalent. In the fact-verification questions with Boolean answers, the judges were instructed to accept responses indicating True, such as  [\textbf{True},\textbf{Yes},\RL{  “صح,”  “نعم”,  “بيان صحيح”}] and those indicating False, such as [\textbf{False}, \textbf{No}, \RL{“خطأ,”  “غير صحيح,”   “بيان كاذب” }]. This rule also allowed long sentence responses to be accepted as True or False as long as the affirmative or negative meaning was clear. The guideline regarding numeric comparisons in terms of precision was that numeric values that matched within a ±0.005 tolerance were considered correct; for example, “1.883” was acceptable as “1.889”. In terms of formatting numeric values, variations were allowed, so “25\%” equated to “0.25,”  and numbers could appear with or without units (e.g., “38000,” “38 \RL {ألف}”). In cases in which the ground-truth answer was a list, such as hotel names, the judges considered two lists as equivalent if they had the same items, without regard to the order, punctuation, or language formatting of these lists; for instance, “\RL{فندق أ و فندق ب}” was considered identical to \RL{[ فندق ب , فندق أ]}.

 Furthermore, if the correct answer included Arabic entity names, such as universities, cities, or streets, only answers given in Arabic were considered correct, as answering in a different language could indicate that the model did not depend on the table as the only source of information; Boolean or numeric answers were exempted from this rule. The rubric also clearly stated that any variations in the punctuation, such as quotes, brackets, and commas; differences in case or whitespace; minor variations in style; and minor spelling mistakes should be ignored. All these guidelines were designed to ensure that our evaluation focused on evaluating the models' ability to understand Arabic tabular data rather than simple formatting differences. Finally, the judges recorded their evaluation as \textbf{TRUE} or \textbf{FALSE} in each answer's evaluation column.

\subsection{LLMs as Judges}\label{subsec:LLMasJudge}
As part of the automated evaluation, two LLMs were used as judges – Qwen \cite{qwen2023} and 4O \cite{openai2024gpt4o} – neither of which had seen the original questions or been used to generate answers. These judge models were provided with the same ground-truth and rubrics as the human judges, and they evaluated the models' answers in two rounds. Initially, each judge evaluated the models' answers independently from the other judge and assigned True or False to each response.\\

\textbf{Assisted Self-Deliberation (ASD) for Disagreement Reduction:}
To increase the agreement between the LLMs as judges and reduce the disagreement with the human evaluation, in the second round, we introduced a self-deliberation mechanism. In this mechanism, (1) we anonymised the judges by giving each one a name: Judge1 for 4O and Judge2 for Qwen; then (2) we used their initial evaluations to generate a disagreement report, highlighting cases in which their evaluations conflicted with those of the other judge. (3) We then gave each judge the report in parts, three cases of conflict at a time, in each case consisting of the model evaluation and the other judge's evaluation. We prompted each judge (model) to justify its own evaluation in light of the rubric and justify the other judge's evaluation by finding which rule supported or opposed each evaluation. In other words, the peer's label played the role of a \textbf{failure flag} that encouraged reflexive self-critique because our self-deliberation protocol asked each judge (4O and Qwen) to revisit its own initial True/False decisions only when they conflicted with the peer's label without ever seeing the peer's reasoning. This ASD mechanism enabled us to avoid coordination costs and reduce the latency that could happen as a result of a two-LLM cross-deliberation loop \cite{liu2024groupdebate,du2023improving}, and to encourage the model to engage in self-criticism. 

\begin{table}[ht!]
\centering
\small
\renewcommand{\arraystretch}{1.1}
\begin{tabular}{|p{0.48\linewidth}|p{0.48\linewidth}|}
\hline
\textbf{4O Self-Deliberation} & \textbf{Qwen Self-Deliberation}\\
\hline
\begin{minipage}[t]{\linewidth}\small
\textbf{Disagreement Analysis (Case 1)}\\[-2pt]
\begin{itemize}
  \item \textbf{Question:} \RL{ما النسبة المئوية للفارق بين الخسارة التقريبية لشركة نخيل العقارية والخسارة الحقيقية؟}
  \item \textbf{Ground Truth:} 4.51\%
  \item \textbf{Model Answer:} 4.50\%
  \item \textbf{Your Original Evaluation:} \textsc{False}
  \item \textbf{Other Judges:} [Judge 2 = \textsc{True}]
\end{itemize}
\end{minipage}
&
\begin{minipage}[t]{\linewidth}\small
\textbf{Disagreement Analysis (Case 1)}\\[-2pt]
\begin{itemize}
  \item \textbf{Question:} \RL{ما النسبة المئوية للفارق بين الخسارة التقريبية لشركة نخيل العقارية والخسارة الحقيقية؟}
  \item \textbf{Ground Truth:} 4.51\%
  \item \textbf{Model Answer:} 4.50\%
  \item \textbf{Your Original Evaluation:} \textsc{True}
  \item \textbf{Other Judges:} [Judge 1 = \textsc{False}]\\
\end{itemize}
\end{minipage}\\
\hline
\begin{minipage}[t]{\linewidth}\small
\textbf{Conflicts with Other Judges}\\[5pt]
Judge 2 marked the answer as \textsc{True}, but the difference is~0.01, which exceeds the $\pm0.005$ tolerance. 
According to \textit{Rule 3} this must be marked as \textsc{False}.\\
\end{minipage}
&
\begin{minipage}[t]{\linewidth}\small
\textbf{Conflicts with Other Judges}\\
 \RL{:لا يوجد تعارض؛ القاعدة 3 تدعم بوضوح التقييم }``\textsc{False}''.
\end{minipage}\\
\hline
\begin{minipage}[t]{\linewidth}\small
\textbf{Rules Supporting My Evaluation}\\[5pt]
   \textbf{Rule 3} (Numeric Match ±$\!$0.005): 4.50\% is \emph{outside} the tolerance; match fails.
\end{minipage}
&
\begin{minipage}[t]{\linewidth}\small
\textbf{Conflicts Against My Evaluation}\\
   \RL{تشترط القاعدة 3 أن يكون الفرق ضمن $\pm0.005$، بينما الفرق الحالي $0.01$   يتعدى الحد.}\\

\end{minipage}\\
\hline
\begin{minipage}[t]{\linewidth}\small
\textbf{Final Decision:} Maintain\\
\textbf{Justification:}\\[5pt]
Although 4.50\% is close to 4.51\%, it lies outside the $\pm0.005$ tolerance, so the answer is \textsc{False}.\\
\end{minipage}
&
\begin{minipage}[t]{\linewidth}\small
\textbf{Final Decision:} Revise\\
\textbf{Justification:}\\

\RL{وفقاً للقاعدة 3، يتعدى الفرق $0.01$ نطاق التسامح ($\pm0.005$)، مما يجعل الإجابة خاطئة} ``\textsc{False}''.
\end{minipage}\\
\hline
\end{tabular}
\caption{Example of self-deliberation output for 4O and Qwen on a numeric-precision disagreement.}
\label{tab:self-deliberation-example}
\end{table}

\section{Results and Discussion}\label{sec:resulta}
\subsection{Analysis of the LLMs' Performance in AraTable QA}

This section presents the empirical results from our evaluation of various LLMs in terms of their ability to understand and extract information from Arabic tabular data in three QA tasks: direct QA, reasoning, and fact verification. In particular, we assessed the performance of Llama 3.3 70B, Mistral Large, DeepSeek-V3, and Jais 70B across 41 Arabic tabular datasets from three distinct sources, as explained in Section \ref{Sec:TabData Collection}. Performance was evaluated in terms of the accuracy between the LLMs' answers and the gold standard ground-truth answer based on human judgement. In other words, humans manually evaluated whether or not the LLMs' answers were correct, as explained in Section \ref{subsec:HumanasJudge}. Accuracy metrics are reported for each model and question type in Tables \ref{tab:accuracy_qtype_Wikipedia} to \ref{tab:accuracy_qtype_LLM} for Wikipedia, real-world, and LLM-generated data sources.

Across all datasets and question types, a clear hierarchy in model performance was observed. DeepSeek-V3 consistently demonstrated the highest accuracy, closely followed by Llama 3.3 70B and Mistral Large. The Jais models exhibited significantly lower performance compared to their counterparts, often struggling particularly with more complex question types. The identical performance between Jais' concise answer and its full answer indicated that the full answers did not provide a distinct advantage in accuracy across all the datasets. A consistent pattern across all models and table sources was the superior performance on direct QA questions, indicating that direct information retrieval from tables is generally easier than tasks requiring inferential capabilities.

Looking in more depth at each of the tables, Table \ref{tab:accuracy_qtype_Wikipedia} shows the performance on Wikipedia tables, revealing that DeepSeek-V3 led in all question categories, achieving 96.15\% on direct QA, 59.23\% on reasoning, and 81.54\% on fact verification. Llama 3.3 70B and Mistral Large also showed strong results, especially on direct QA questions. Jais and Jais full-answer models exhibited the lowest accuracy, with 30\% for reasoning and 63.85\% for direct QA, suggesting challenges in processing Wikipedia-style tabular data. Table \ref{tab:accuracy_qtype_Realworld} presents the results for real-world tables, which are often more heterogeneous and less structured than Wikipedia tables. DeepSeek maintained its lead, excelling on direct QA questions with 98\% accuracy and performing well on fact verification with 80\% accuracy, but its accuracy dropped to 48\% in reasoning questions. Llama's reasoning accuracy significantly dropped to 20\% on this dataset. The Jais models continued to struggle, showing very low accuracy across all question types, particularly reasoning with 14\%. Real-world tables proved to be more challenging for reasoning tasks for most models. Table \ref{tab:accuracy_qtype_LLM}, displaying the results for LLM-generated tables, provides interesting insights. DeepSeek-V3 achieved a perfect score of 100\% on direct QA questions, demonstrating exceptional capability in extracting direct information from synthetic tables. Both DeepSeek-V3 and Mistral Large performed identically on reasoning questions (48\%), which is a notable improvement for Mistral Large compared to real-world data. Llama 3.3 70B also showed improved performance in direct QA and fact verification. The Jais models saw a slight increase in direct QA accuracy (56\%) but remained significantly behind the other models.

The most striking observation was the consistent gap between direct QA questions and other question types. Direct QA questions, which primarily involved direct extraction of information, consistently yielded high accuracies (often above 90\% for top models). In contrast, reasoning questions consistently posed the greatest challenge, with accuracies generally below 60\% even for the best-performing models. This highlights a fundamental limitation of current LLMs in performing complex logical inferences using tabular data, especially when the data is not perfectly structured or contains subtle nuances. Fact verification questions fell in between, suggesting that they require a moderate level of understanding beyond simple extraction but less complex inferential steps than reasoning. In addition, DeepSeek's clear lead, followed closely by Llama 3.3 70B and Mistral Large, hints that larger token budgets and more diverse pre-training corpora (especially those that include structured data and Arabic sources) matter more than architecture tweaks alone. Although Jais 70B is Arabic-centric, it lags badly on reasoning, suggesting that raw Arabic coverage is not enough; exposure to tabular reasoning patterns during pre-training/fine-tuning is likely required. These findings provide valuable insights into the current state of LLM performance in Arabic tabular data and identify key areas for future research and development, particularly to improve reasoning capabilities and robustness across diverse, real-world datasets.

\begin{table}[ht!]
\centering
\small
\renewcommand{\arraystretch}{1.1}
\begin{tabular}{lcccccc}
\toprule
Question Type & Llama70B & DeepSeek-v3 & Mistral-L & Jais70B & Jais Full answer\\
\midrule

Direct QA & 0.90 & 0.96 & 0.92 & 0.63 & 0.63\\
Fact Verification & 0.75 & 0.81 & 0.70 & 0.53 & 0.53\\
Reasoning & 0.45 & 0.59 & 0.53 & 0.30 & 0.30\\
\bottomrule
Overall&0.71 & 0.79 & 0.72 & 0.49 & 0.49\\
\bottomrule
\end{tabular}
\caption{Accuracy per Model by Question Type for Wikipedia tables}
\label{tab:accuracy_qtype_Wikipedia}
\end{table}

\begin{table}[ht!]
\centering
\small
\renewcommand{\arraystretch}{1.1}
\begin{tabular}{lcccccc}
\toprule
Question Type & Llama70B & DeepSeek-v3 & Mistral-L & Jais70B & Jais Full answer\\
\midrule
Direct QA & 0.86 & 0.98 & 0.90 & 0.40 & 0.40\\
Fact Verification & 0.74 & 0.80 & 0.72 & 0.38 & 0.38\\
Reasoning & 0.20 & 0.48 & 0.30 & 0.14 & 0.14\\
\bottomrule
Overall&0.60 & 0.75 & 0.64 & 0.31 & 0.31\\
\bottomrule
\end{tabular}
\caption{Accuracy per Model by Question Type for the considered Real-World tables}
\label{tab:accuracy_qtype_Realworld}
\end{table}

\begin{table}[h!]
\centering
\small
\renewcommand{\arraystretch}{1.1}
\begin{tabular}{lcccccc}
\toprule 
Question Type & Llama70B & DeepSeek-v3 & Mistral-L & Jais70B & Jais Full answer
\\
\midrule
Direct QA & 0.92 & 1.00 & 0.96 & 0.56 & 0.56\\
Fact Verification & 0.76 & 0.76 & 0.60 & 0.60 & 0.60 \\
Reasoning & 0.32 & 0.48 & 0.48 & 0.20 & 0.20 \\
\bottomrule
Overall&0.67 & 0.75 & 0.68 & 0.45 & 0.45\\
\bottomrule
\end{tabular}

\caption{Accuracy per Model by Question Type for the LLM-generated tables}
\label{tab:accuracy_qtype_LLM}
\end{table}

\subsection{LLMs as Judges: An Analysis of Automatic Evaluation Performance}

This section analyses the performance of the LLMs, specifically Qwen and 4O, when utilised as automatic evaluators (or judges) to assess other LLMs' performance as explained in Section\ref{subsec:LLMasJudge}. Their accuracy was compared to human evaluators across all the considered datasets in Tables \ref{tab:wikipedia_pre} to \ref{tab:llmgen_pre} for Wikipedia, real-world, and LLM-generated datasets; the evaluation assessed the judgement accuracy both before and after a deliberation phase, with a particular focus on the signed gap in accuracy relative to the human judgement baseline.

Before deliberation, Qwen's performance was closely aligned with human evaluators for higher-performance LLMs such as Llama 3.3 70B  (e.g., $\Delta$= \textminus{}0.01 on real-world), DeepSeek-V3, and Mistral Large across all datasets, with very small differences in precision. However, for models such as Jais70B and Jais-Full, Qwen often displayed a slight positive bias, meaning it tended to rate these models slightly higher than human evaluators did (e.g., Wikipedia Jais-Full $\Delta$=+0.03, real-world Jais70B $\Delta$=+0.07). ). On the LLM-generated dataset, Qwen even showed a higher rate than human judges for some models (e.g., Mistral Large $\Delta$=+0.04, Jais70B $\Delta$=+0.08).  In contrast, the 4O model consistently underrating the performance of target LLMs compared to human evaluators in almost all target LLMs and datasets. It exhibited significant negative gaps in accuracy (e.g., Wikipedia Jais-Full $\Delta$= \textminus{}0.15, real-world Jais-Full $\Delta$= \textminus{}0.16, real-world DeepSeek $\Delta$= \textminus{}0.07). This indicates that 4O's judgements were often less accurate and less aligned with human assessments in the initial, pre-deliberation stage.

\begin{figure}[ht!]
  \centering
  \begin{subfigure}[b]{0.30\textwidth}%
    \includegraphics[width=\linewidth]{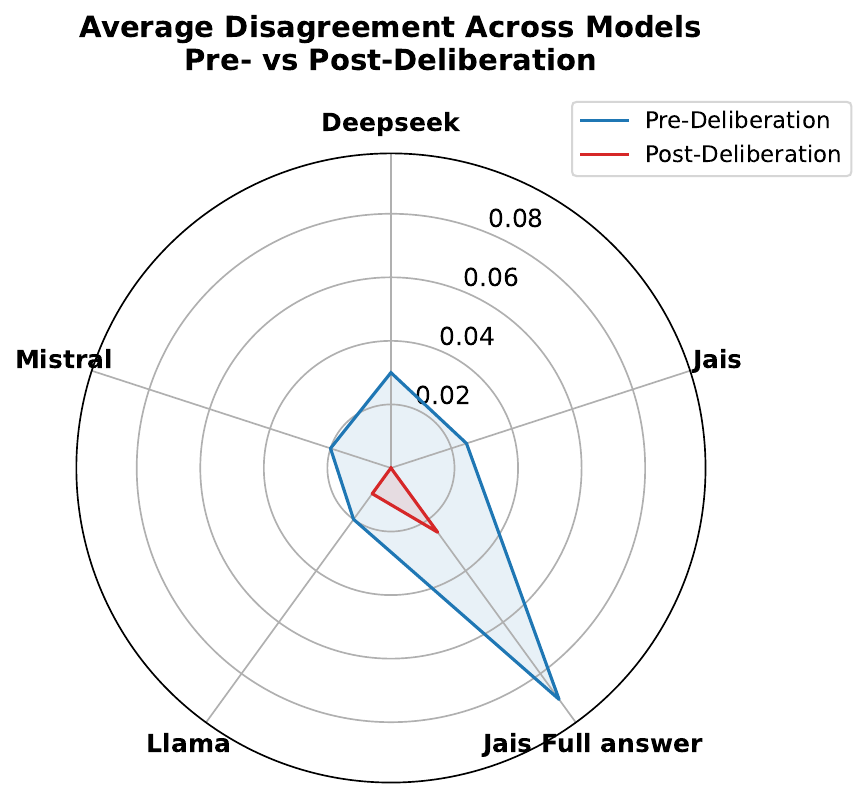}%
    \caption{Wikipedia Dataset}%
    \label{fig:Wikipedia}%
  \end{subfigure}%
  \hfill%
  \begin{subfigure}[b]{0.30\textwidth}
    \includegraphics[width=\linewidth]{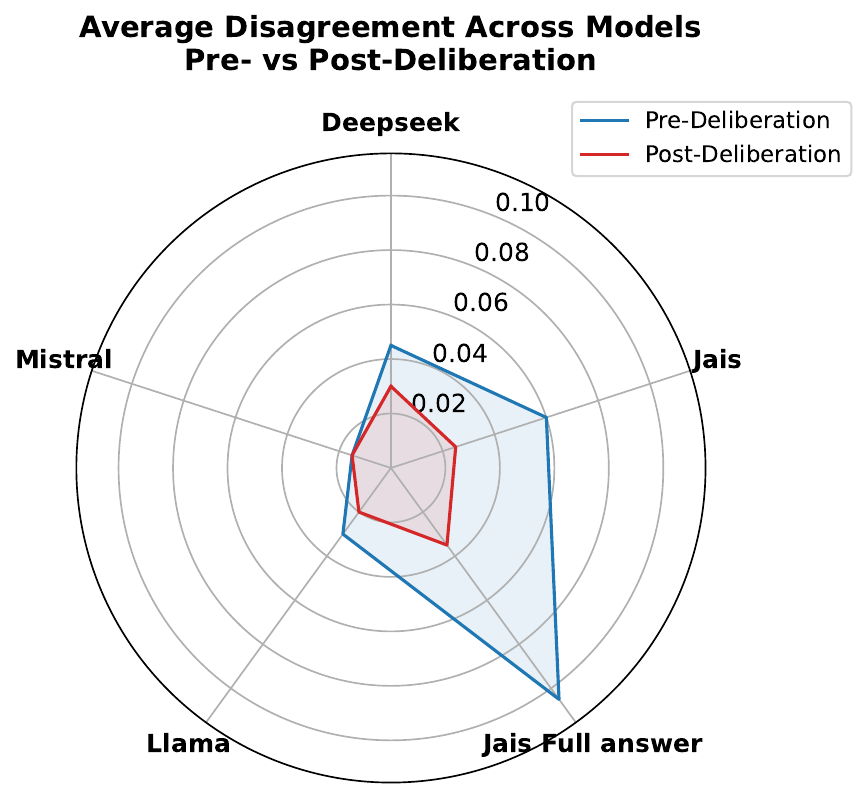}
    \caption{Real-world Dataset}
    \label{fig:RealWorld}
  \end{subfigure}%
  \hfill%
  \begin{subfigure}[b]{0.30\textwidth}
    \includegraphics[width=\linewidth]{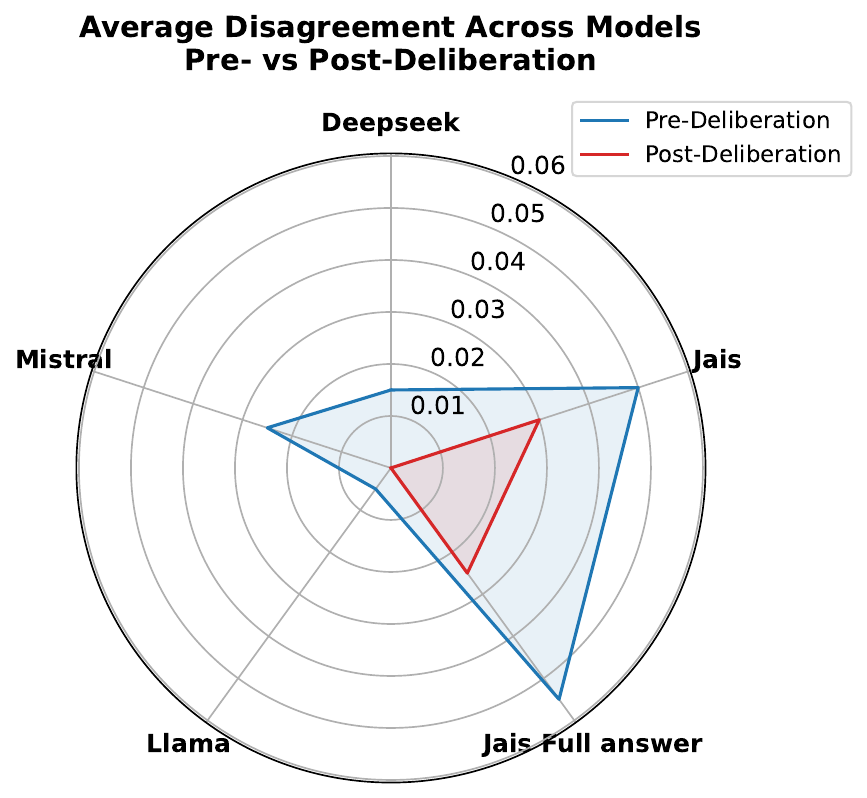}
    \caption{ LLM-generated Dataset}
    \label{fig:LLMgenerated}
  \end{subfigure}

  \caption{The impact of Deliberation Phase on reducing the disagreement between Human and LLMs raters.}
  \label{fig:authors}
\end{figure}

Following the deliberation phase, a remarkable improvement was observed in Qwen's performance as an LLM judge. Its judgement accuracy consistently converged with human baselines across all datasets (Wikipedia, real-world, and LLM-generated), frequently achieving a perfect match for various target LLMs, including Llama70B, DeepSeek-V3, Mistral Large, Jais70B, and Jais-Full (e.g., real-world Llama70B $\Delta$=+0.00, Wikipedia Mistral Large  $\Delta$=+0.00). This transformative effect highlights the deliberation mechanism's effectiveness in refining Qwen's internal evaluation criteria, allowing it to accurately mirror human assessment. For instance, while Qwen sometimes exhibited slight discrepancies before deliberation (e.g., a positive bias towards Jais models on the real-world dataset), these were consistently eliminated post deliberation, bringing its automated judgements into full congruence with human standards. In contrast, 4O's improvement was more limited; while it showed some gains, its alignment with human evaluators remained less consistent compared to Qwen's. It often retained negative gaps, such as for Jais-Full on the Wikipedia ($\Delta$= \textminus{}0.05) and real-world ($\Delta$= \textminus{}0.05) datasets, suggesting that 4O either benefited less thoroughly from the deliberation process or that its inherent biases were more resistant to correction. Figure\ref{fig:authors} shows the average unsigned error (mean $|$ $\Delta$ $|$) between the accuracy of the LLM raters and the human baseline, both before (blue polygons) and after (red polygons) the ASD step across all the three datasets: (a) Wikipedia, (b) real-world, and (c) LLM-generated questions. Self-deliberation consistently and substantially reduced the gap between the accuracy of the LLM raters and the human raters for all models, as evidenced by the red shape in each subplot being nested well inside the blue one. The reduction is most noticeable in Jais' full answer, which indicates that self-deliberation is most important when the LLM judges deal with models that suffer from verbosity. Overall, the figure confirms that ASD roughly halves the average difference with human judgements.

In summary, the deliberation phase proved to be a critical component for enhancing the reliability and human alignment of LLMs acting as judges. In Table \ref{tab:self-deliberation-example}, we show a case in which each model decides to revise or maintain its evaluation after the ASD phase. Qwen, in particular, achieves near-perfect performance compared to human baselines after deliberation, highlighting its strong potential as a highly accurate automatic evaluator, significantly advancing the feasibility of using AI for robust assessment. While 4O's results indicate some variability in the effectiveness of deliberation depending on the specific LLM judge, the overall impact of this phase represents a major step forward in developing more dependable and human-aligned automated evaluation methodologies. Even if we assign the final decision to one model, consulting a second model remains valuable because its alternative output encourages diverse reasoning and helps the main model to think more critically through the ASD mechanism.

\begin{table*}[ht!]
  \centering
  \small
  \setlength{\tabcolsep}{1.5pt}
  \begin{tabular}{l
                  S S S S S
                  S[table-format = +1.2]
                  S[table-format = +1.2]
                  S[table-format = +1.2]
                  S[table-format = +1.2]
                  S[table-format = +1.2]
                  }
    \toprule
      & \multicolumn{5}{c}{\textbf{Accuracy ($\uparrow$)}} &
        \multicolumn{5}{c}{$\Delta$ vs Human} \\
    \cmidrule(lr){2-6}\cmidrule(lr){7-11}
    \textbf{Evaluator}  &
      {Llama} & {DeepSeek} & {Mistral} & {Jais} & {Jais-Full} &
      {$\Delta_{\text{L70B}}$} & {$\Delta_{\text{DS}}$} &
      {$\Delta_{\text{Mis}}$}  & {$\Delta_{\text{J70B}}$} &
      {$\Delta_{\text{J-F}}$} \\
    \midrule
    Human Raters & 0.71 & 0.79 & 0.72 & 0.49 & 0.49 & {+0.00} & {+0.00} & {+0.00} & {+0.00} &{+0.00} \\
    Qwen before ASD         & 0.69 & 0.77 & 0.71 & 0.51 & 0.52 & {-0.02} & {-0.02} & {-0.01} & {+0.02} & {+0.03} \\
    Qwen after ASD       & 0.71 & 0.79 & 0.72 & 0.49 & 0.49 & {+0.00} & {+0.0}0 & {+0.00} & {+0.00} &{ +0.00} \\
    4O before ASD         & 0.69 & 0.75 & 0.69 & 0.46 & 0.34 & {-0.02} & {-0.04} & {-0.03} & {-0.03} & {-0.15} \\
     4O after ASD          & 0.69 & 0.79 & 0.72 & 0.49 & 0.44 &{ -0.02} & {+0.00} & {+0.00} & {+0.00} & {-0.05} \\
    \bottomrule
  \end{tabular}
    \caption{Wikipedia dataset accuracies and signed gaps to the Human baseline reported \textbf{before and after the assisted self-deliberation (ASD)}.}\label{tab:wikipedia_pre}
\end{table*}

\begin{table*}[ht!]
  \centering
  
  \label{tab:pre_delib}
  \small
  \setlength{\tabcolsep}{1.5pt} 
  \begin{tabular}{
      l
      S S S S S
      S[table-format = +1.2]
      S[table-format = +1.2]
      S[table-format = +1.2]
      S[table-format = +1.2]
      S[table-format = +1.2]
  }
    \toprule
      & \multicolumn{5}{c}{\textbf{Accuracy ($\uparrow$)}} &
        \multicolumn{5}{c}{$\Delta$ vs Human} \\
    \cmidrule(lr){2-6}\cmidrule(lr){7-11}
    \textbf{Evaluator}
      & {Llama} & {DeepSeek} & \textbf{Mistral}
      & {Jais}  & {Jais-Full}
      & {$\Delta_{\text{L70B}}$}
      & {$\Delta_{\text{DS}}$}
      & {$\Delta_{\text{Mis}}$}
      & {$\Delta_{\text{J70B}}$}
      & {$\Delta_{\text{J-F}}$} \\
    \midrule
    Human & 0.60 & 0.75 & 0.64 & 0.31 & 0.31  & {+0.00} & {+0.00} & {+0.00} & {+0.00} & {+0.00} \\
    Qwen before ASD & 0.59 & 0.73 & 0.64 & 0.38 & 0.36  & {-0.01} & {-0.02} & {+0.00} & {+0.07} & {+0.05} \\
     Qwen after ASD & 0.60 & 0.75 & 0.64 & 0.31 & 0.31  & {+0.00} & {+0.00} & {+0.00} & {+0.00} & {+0.00} \\
    4O before ASD   & 0.55 & 0.68 & 0.61 & 0.26 & 0.15  & {-0.05} & {-0.07} & {-0.03} & {-0.05} & {-0.16} \\
    
    4O after ASD    & 0.58 & 0.75 & 0.64 & 0.31 & {0.26}  & {-0.02} & {+0.00} &{ +0.00} & {+0.00} & {-0.05} \\
    \bottomrule
  \end{tabular}
  \caption{ Real-world dataset accuracies and signed gaps to the Human baseline reported \textbf{before and after the assisted self-deliberation (ASD)}.}
\end{table*}

\begin{table*}[ht!]
  \centering

  \small
  \setlength{\tabcolsep}{1.5pt}
  \begin{tabular}{l
                  S S S S S
                  S[table-format = +1.2]
                  S[table-format = +1.2]
                  S[table-format = +1.2]
                  S[table-format = +1.2]
                  S[table-format = +1.2]}
    \toprule
      & \multicolumn{5}{c}{\textbf{Accuracy ($\uparrow$)}} &
        \multicolumn{5}{c}{$\Delta$ vs Human} \\
    \cmidrule(lr){2-6}\cmidrule(lr){7-11}
    \textbf{Evaluator}  &
      {Llama} & {DeepSeek} & {Mistral} &
      {Jais}  & {Jais-Full} &
      {$\Delta_{\text{L70B}}$} & {$\Delta_{\text{DS}}$} &
      {$\Delta_{\text{Mis}}$}  & {$\Delta_{\text{J70B}}$} &
      {$\Delta_{\text{J-F}}$} \\
    \midrule
    Human Raters & 0.67 & 0.75 & 0.68 & 0.45 & 0.45 & {+0.00} & {+0.00} & {+0.00} & {+0.00} & {+0.00 }\\
    Qwen befor ASD  & 0.68 & 0.76 & 0.72 & 0.53 & 0.52 & {+0.01} & {+0.01} & {+0.04} & {+0.08 }& {+0.07} \\
    Qwen after ASD        & 0.67 & 0.75 & 0.68 & 0.49 & 0.48 & {+0.00} & {+0.00} & {+0.00} & {+0.04} & {+0.03} \\
    4O before ASD  & 0.67 & 0.73 & 0.69 & 0.47 & 0.41 & {+0.00} & {-0.02} & {+0.01} & {+0.02} & {-0.04} \\

    4O after ASD          & 0.67 & 0.75 & 0.68 & 0.47 & 0.47 &{ +0.00} & {+0.00} & {+0.00} & {+0.02} & {+0.02} \\
    \bottomrule
  \end{tabular}
    \caption{LLM-generated dataset accuracies and signed gaps to the Human baseline reported \textbf{before and after the assisted self-deliberation (ASD)}.}\label{tab:llmgen_pre}
\end{table*}

\section{Conclusion and Future Work}\label{sec:Concl}
In this paper, we presented \textbf{AraTable}, a question answering benchmark for use with Arabic tabular data. The benchmark provides a framework for evaluating LLMs on three distinct QA tasks: direct extraction, reasoning, and fact verification. Each question type was designed to probe different aspects of model understanding and cognitive reasoning capabilities. Our evaluation highlights that while some models perform well on surface-level tasks, cognitive reasoning and verification remain challenging, revealing significant gaps in Arabic tabular understanding. Although this paper focuses on zero-shot QA evaluation, AraTable can also serve as a foundation for fine-tuning or adapting models to Arabic tabular reasoning tasks.

Building on this foundation, future work will explore the few-shot and fine-tuned settings in Arabic tabular data, particularly for complex reasoning scenarios. Extending the benchmark to include larger and more diverse tables with varied formats, such as multi-table or hierarchical structures, and to explore efficient methods for handling such data, including retrieval-augmented generation, is another promising direction. We hope this work encourages future efforts towards the advancement of Arabic table understanding in LLMs.

\section{Limitations}\label{sec:limit}
Our evaluation is associated with several limitations. First, we tested only a limited number of models, focusing exclusively on open-source models that support the Arabic language. While these models have shown strong performance across various tasks on Arabic benchmarks, some models may be more optimal for this particular type of reasoning using tabular data. Second, in our evaluation, LLMs were evaluated exclusively in a zero-shot context. The models' performance may be improved with more examples or a few-shot context, which we leave for future research. Third, our assessment was focused primarily on small tables, which contained only 40 rows each. This focus was primarily due to the limitations of certain models in handling large data volumes.

\section{Ethical Statement}
All data included in AraTable were collected from publicly available sources such as Arabic Wikipedia, government portals, and open data platforms, all under non-restrictive licenses. We preserved original table content without introducing modifications and credited each source accordingly (see Tables\ref{tab:Wikitable}\&\ref{tab:Real-worldtable}). Care was taken to exclude tables or generated QA pairs containing offensive, sensitive, or controversial content. While the LLMs used in the QA generation and evaluation phases may have exhibited biases, all model outputs were manually reviewed to ensure that no harmful, discriminatory, or inappropriate content was present in the released benchmark.

\bibliography{sn-bibliography}

\end{document}